\documentclass{article}

\usepackage{PRIMEarxiv}

\usepackage[utf8]{inputenc} 
\usepackage[T1]{fontenc}    
\usepackage{hyperref}       
\usepackage{url}            
\usepackage{booktabs}       
\usepackage{amsfonts}       
\usepackage{nicefrac}       
\usepackage{microtype}      
\usepackage{lipsum}
\usepackage{fancyhdr}       
\usepackage{graphicx}       
\graphicspath{{media/}}     
\usepackage{tcolorbox}
\usepackage{float}
\title{Evaluating the Clinical Impact of Generative Inpainting on Bone Age Estimation 

}

\author{
  Felipe Akio Matsuoka\thanks{First author} \\
  Universidade Federal de São Paulo \\
  Dasa \\
  \texttt{felipe.matsuoka.ext@dasa.com.br} \\
  \And
  Eduardo M. J. M. Farina, MD \\
  Universidade Federal de São Paulo \\
  Dasa \\
  \And
  Augusto Sarquis Serpa, MD \\
  Universidade Federal de São Paulo \\
  Universidade de São Paulo \\
  \And
  Soraya Monteiro, MD, PhD \\
  Universidade Federal de São Paulo \\
  \And
  Rodrigo Ragazzini, MD, PhD \\
  Universidade Federal de São Paulo  \\
  \And
  Nitamar Abdala, MD, PhD \\
  Universidade Federal de São Paulo  \\
  \And
  Marcelo Straus Takahashi, MD, PhD \\
  University of North Carolina at Chapel Hill \\
  Dasa \\
  \And
  Felipe Campos Kitamura, MD, PhD \\
  Universidade Federal de São Paulo \\
  Bunkerhill Health \\
}

\begin{document}
\maketitle

\begin{abstract}

Generative foundation models can remove visual artifacts through realistic image inpainting, but their impact on medical AI performance remains uncertain. Pediatric hand radiographs often contain non-anatomical markers, and it is unclear whether inpainting these regions preserves features needed for bone age and gender prediction. To evaluate the clinical reliability of generative model–based inpainting for artifact removal, we used the RSNA Bone Age Challenge dataset, selecting 200 original radiographs and generating 600 inpainted versions with \texttt{gpt-image-1} using natural language prompts to target non-anatomical artifacts. Downstream performance was assessed with deep learning ensembles for bone age estimation and gender classification, using mean absolute error (MAE) and area under the ROC curve (AUC) as metrics, and pixel intensity distributions to detect structural alterations. Inpainting markedly degraded model performance: bone age MAE increased from 6.26 to 30.11 months, and gender classification AUC decreased from 0.955 to 0.704. Inpainted images displayed pixel-intensity shifts and inconsistencies, indicating structural modifications not corrected by simple calibration. These findings show that, although visually realistic, foundation model–based inpainting can obscure subtle but clinically relevant features and introduce latent bias even when edits are confined to non-diagnostic regions, underscoring the need for rigorous, task-specific validation before integrating such generative tools into clinical AI workflows.

\end{abstract}


\newpage

\section{Introduction}
Generative vision–language models like  \texttt{gpt-image-1} \cite{openai_gpt_image_1} have demonstrated strong capabilities in interpreting and editing complex visual scenarios\cite{roledeeplearninginmedicalimage}. Their applications span image captioning, reasoning, and image inpainting, filling in missing regions with visually plausible content. While recent work has applied diffusion-based inpainting for general tasks and even generating synthetic MRIs with promising anatomical coherence \cite{lugmayr2022repaintinpaintingusingdenoising, rouzrokh2023multitaskbraintumorinpainting}, these approaches rely on explicit masks to determine the edit area. However, it remains unknown whether foundation models such as  \texttt{gpt-image-1}, which have shown an ability to reason about complex visual contexts, can achieve high-quality edits using only natural language prompts without requiring users to highlight a specific region of interest. These prior efforts have also largely emphasized generation quality rather than clinical validation, leaving unanswered whether such outputs preserve medically relevant features \cite{article}.
This gap is particularly relevant in radiology, where even subtle anatomical details carry diagnostic value. As foundation models gain popularity for many downstream tasks, a key challenge emerges: their versatility and scale often come at the cost of verifiability, interpretability, and domain-specific validation \cite{khan2025comprehensivesurveyfoundationmodels}. In this context, there is a need to understand whether their inpainted outputs preserve the clinical value of the original images.
In this study, we investigate the reliability of \texttt{gpt-image-1} in inpainting pediatric hand radiographs. Specifically, we use it to remove burned-in artifacts, such as labels, borders, and film annotations, and reconstruct the underlying anatomy. Rather than optimizing performance, our objective is to validate the generative model’s output: are the inpainted images anatomically similar, and do they support consistent clinical inference?
The results aim to inform future applications of foundation models in dataset refinement, synthetic augmentation, and artifact removal workflows, while highlighting the limitations and failure modes of inpainting in medically sensitive contexts.

\section{Methods}

We conducted a retrospective, comparative analysis using a matched case--control--like design to evaluate the impact of generative inpainting on the performance of deep learning models for two tasks: bone age estimation and gender classification from pediatric hand radiographs. The study was based on the publicly available RSNA Bone Age Challenge dataset \cite{halabi2019rsna}, which includes radiographs of patients aged 0 to 19 years, each labeled with ground truth bone age and gender by expert radiologists. While the dataset was originally released for a bone age estimation challenge, our study differs by investigating how generative inpainting affects downstream AI model performance.

For this analysis, we used the complete test subset from the RSNA dataset comprising 200 pediatric hand radiographs. Each original radiograph served as a control. To generate the corresponding case images, we created three synthetic inpainted versions per patient ID using OpenAI’s \texttt{gpt-image-1} to digitally remove non-anatomical artifacts and extraneous labels located outside the hand region, such as laterality markers or institutional identifiers. This was achieved through OpenAI’s image generation API, guided by the following prompt:
\vspace{15pt} 
\begin{tcolorbox}[colback=gray!5,colframe=black,title={Inpainting Prompt}]
\textit{``Enhance this pediatric hand X-ray by digitally removing non-anatomical artifacts or labels outside the hand region, while preserving all anatomical details and ensuring the radiograph remains realistic and diagnostically useful.''}
\end{tcolorbox}
\vspace{15pt} 

Original images were resized to 1024\,$\times$\,1024 pixels and submitted to the \texttt{gpt-image-1} model with the  quality parameter being set to ``high'', alongside a full white mask to define the editable region. It is worth mentioning that in this API, the quality parameter controls rendering fidelity rather than spatial resolution—higher settings increase the model’s compute per pixel, resulting in finer detail and fewer artifacts at the cost of longer generation time. No preprocessing was applied before inpainting. This was an intentional choice to evaluate the generative model’s performance in an ``out-of-the-box'' setting, reflecting common use cases in which foundation models are applied to raw imaging data without modality-specific adjustments.

All radiographs were stored as PNG files with varying spatial dimensions and no embedded metadata. The training set (n=12{,}611) had a mean chronological age of 127.3 months (SD 41.2, median 132.0), with 54.2\% male and 45.8\% female. The test set (n=200) had a mean age of 132.1 months (SD 43.1, median 139.4), with a balanced gender distribution (50\% male, 50\% female).

\newpage
\subsection{Models Used For Validation}
\subsubsection{Bone Age Predictor}
Bone age estimation was performed using a sex-specific ensemble of ResNet50 \cite{7780459} models, implemented in PyTorch (v2.7.1). Separate ensembles were trained for male and female subjects using the RSNA Bone Age Challenge dataset, initialized with ImageNet-pretrained weights \cite{5206848}. Each ensemble consisted of multiple models trained with five-fold cross-validation, and predictions during inference were computed by averaging outputs from the models corresponding to the subject’s sex. Radiographs, originally in grayscale, were converted to RGB via channel duplication. Preprocessing was performed using MONAI \cite{cardoso2022monaiopensourceframeworkdeep} and Albumentations Python packages and included automatic foreground cropping (nonzero-pixel bounding boxes to isolate the hand), resizing to 320×320 pixels, and intensity normalization. Data augmentation during training included random affine transformations (rotation ±15°, translation, scaling), contrast/brightness shifts, and sharpening. Training was conducted for 50 epochs using the Adam optimizer \cite{kingma2017adammethodstochasticoptimization} with gradient accumulation (steps = 10) and cosine annealing learning rate scheduling. The loss function was Smooth L1 (Huber loss)\cite{gokcesu2021generalizedhuberlossrobust}, and optional gender embeddings were used in some configurations via an auxiliary linear layer. All predictions were continuous bone age estimates in months.

\subsubsection{Gender Predictor}
Gender classification was performed using an ensemble of five ResNet50 models, implemented using PyTorch. All models were trained on the RSNA Bone Age Challenge training dataset using grayscale radiographs, which were converted to RGB by channel duplication. Preprocessing followed the same pipeline as the bone age task, including foreground cropping, resizing to 320×320 pixels, spatial padding, and intensity normalization to 0-1.
Training was conducted on stratified 90/10 train/validation splits, using the Adam optimizer (learning rate = 1e-4), binary cross-entropy loss with logits, and a batch size of 32, for five epochs. Data augmentation was applied during training using Albumentations, which included affine transformations (shift, scale, rotation), brightness/contrast adjustments, and stochastic blurring. In inference, final predictions were obtained by averaging the sigmoid outputs across the five models

\subsection{Statistical Analysis}

Bone age model performance was evaluated using mean absolute error (MAE), root mean squared error (RMSE), and mean bias, calculated separately for original and inpainted images. To evaluate whether inpainting introduced a systematic, correctable bias in bone age estimation, we performed a linear calibration analysis. For each study in the inpainted group, predicted bone age values were regressed against the corresponding ground truth labels using ordinary least squares. The fitted slope and intercept defined a calibration curve, which was then applied to the model’s raw predictions. Performance metrics—including MAE, RMSE, and mean bias—were recomputed before and after calibration to quantify improvements and determine whether discrepancies reflected a linear offset (e.g., global under- or overestimation) or deeper nonlinear distortions. This approach allowed us to assess whether the impact of inpainting on downstream predictions followed a consistent proportional trend amenable to simple correction.
For the gender classification ensemble, we additionally computed the area under the receiver operating characteristic curve (AUC) to evaluate classification performance. To further characterize classification errors, confusion matrices were generated for both original images and inpainted images. Notice that for each patient, the final prediction on synthetic data was determined by majority voting across the three images. To assess the consistency of the inpainting process itself, we computed per-patient statistics across the three generated images. For bone age, we calculated the standard deviation, and for gender, we quantified the frequency of prediction inconsistencies, defined as cases in which not all three predictions agreed.
Finally, we computed and compared the pixel intensity distributions of the original and inpainted images to quantify grayscale shifts introduced by the generative process. Density plots were used to visualize the relative frequency of intensity values (0–255), and changes in histogram shape were interpreted as potential indicators of contrast loss or textural smoothing.

\section{Results}
\begin{figure}[ht]
    \centering
    \includegraphics[width=0.85\linewidth]{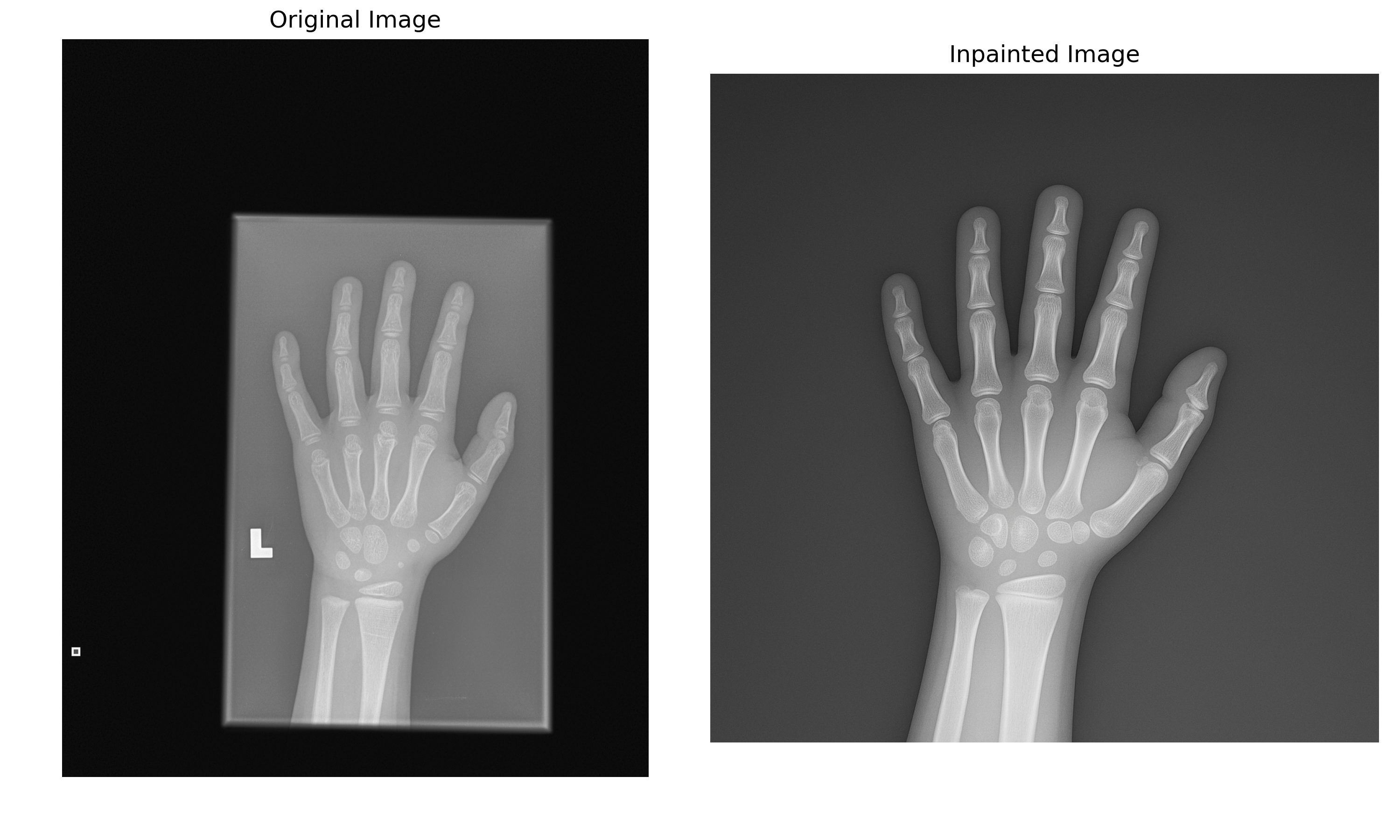}
    \caption{Comparison of original and inpainted pediatric hand X-ray. The image generator not only removed artifacts but also made the bones appear more mature.}
    \label{fig:fig1}
\end{figure}

All 200 selected radiographs were successfully processed, and no images were excluded or failed during inference. In total, the process yielded a paired dataset of 800 images—200 original and 600 inpainted versions (Figure ~\ref{fig:fig1}).

For the bone age estimation task, the ensemble achieved a mean absolute error (MAE) of 6.26 months (95\% CI, 5.60–6.89) and a root mean squared error (RMSE) of 7.79 months (95\% CI, 7.02–8.65) on the original test set. Performance deteriorated markedly on the inpainted images, with an MAE of 30.11 months (95\% CI, 28.45–34.90) and an RMSE of 39.27 months (95\% CI, 36.06–42.94). Predictions displayed a large positive bias of +28.41 months, consistent with systematic overestimation.

\begin{figure}[H]
    \centering
    \includegraphics[width=0.95\linewidth]{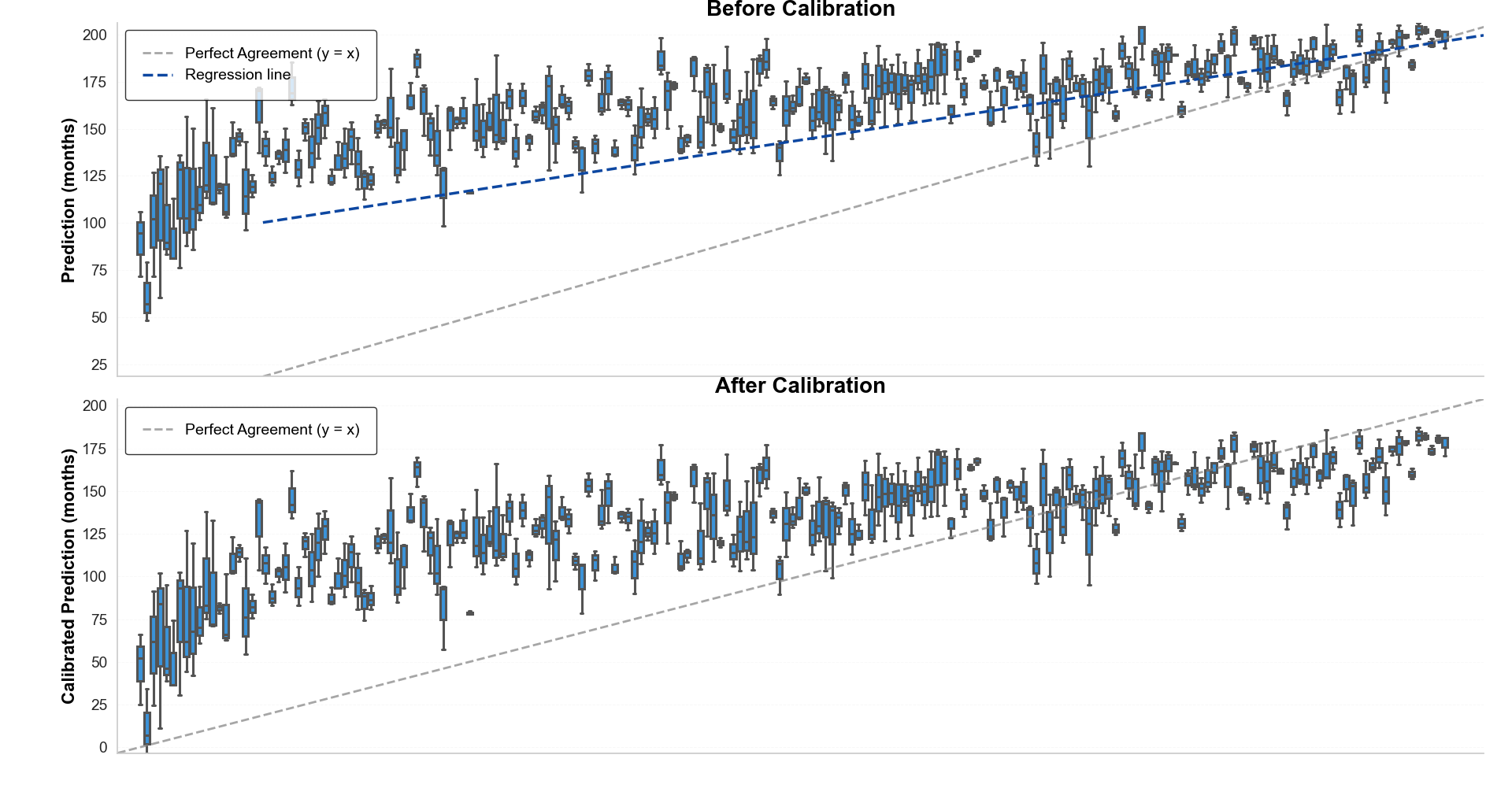}
    \caption{Model predictions on original vs.\ inpainted images (top: scatter with regression and identity line). Right: calibrated predictions vs.\ ground truth. Dashed lines indicate perfect agreement.}
    \label{fig:fig2}
\end{figure}

Applying the linear calibration model removed this mean bias; however, substantial residual error persisted (MAE 19.55 months, RMSE 25.00 months). These results indicate that the degradation introduced by inpainting cannot be explained by a simple linear shift or scaling, but instead reflects deeper nonlinear disruptions in features essential for age estimation (Figure ~\ref{fig:fig2}).

\begin{figure}[ht]
    \centering
    \includegraphics[width=\linewidth]{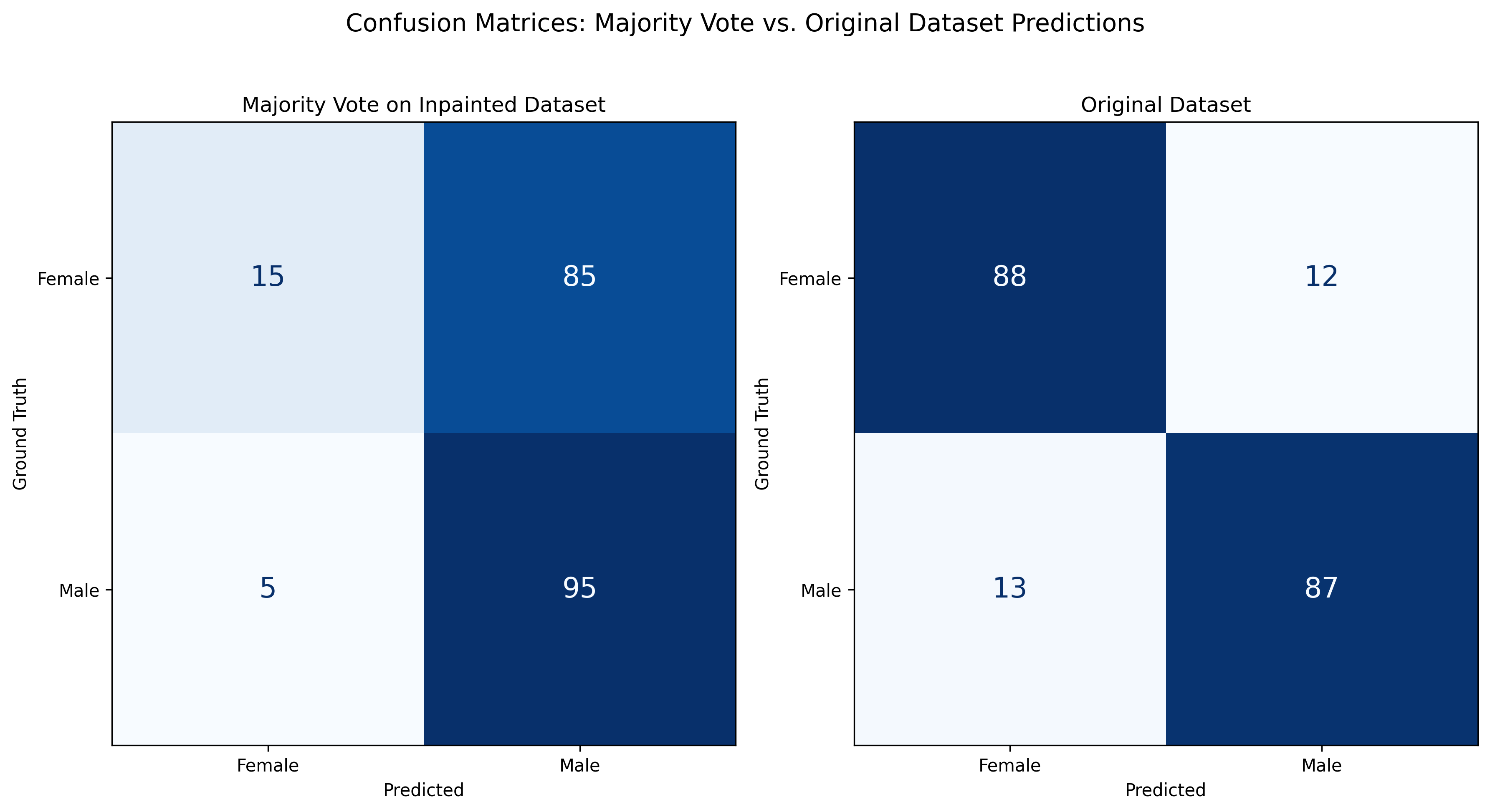}
    \caption{Confusion matrices for gender classification on the inpainted dataset and on the original dataset.}
    \label{fig:fig3}
\end{figure}

For the gender classification task, model performance also worsened after inpainting. On the original test set, the ensemble achieved an AUC of 0.956 (95\% CI, 0.925–0.979). In contrast, the AUC dropped to 0.704 (95\% CI, 0.631 - 0.780) on the inpainted images. Further insight was gained from the confusion matrices comparing prediction errors across the original and inpainted datasets (Figure ~\ref{fig:fig3}). On the original radiographs, misclassifications were balanced (12 female-to-male, 13 male-to-female). However, after inpainting, the model exhibited a substantial shift in error distribution: false male classifications increased, with 85 female radiographs incorrectly labeled as male.

To assess the consistency of the image generation process, we analyzed per-patient variability in model predictions across the three generated images. For bone age, the standard deviation of predictions within each patient averaged 8.77 months (±5.81), with intra-patient variability ranging from 0.19 to 32.47 months.  For gender classification, 29.5\% of patients had at least one conflicting prediction across the three images, indicating notable instability in the model’s outputs across synthetic variations.

\begin{figure}[ht]
    \centering
    \includegraphics[width=0.95\linewidth]{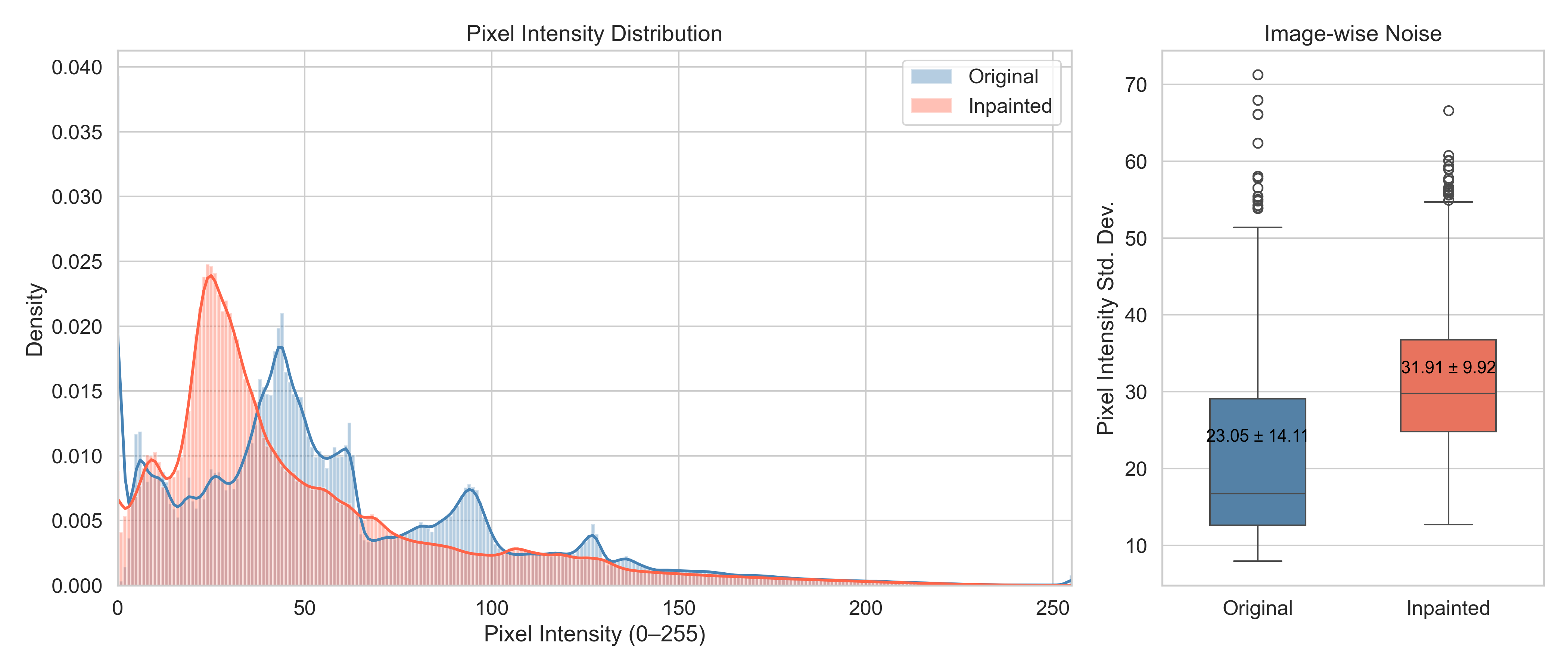}
    \caption{Visual and quantitative comparison of image appearance before and after inpainting. (Left) Pixel intensity distribution of original and inpainted hand radiographs, showing the normalized frequency of grayscale values (0--255) across all images. (Right) Boxplot of per-image pixel intensity standard deviation, confirming a significant decrease in noise after inpainting.}
    \label{fig:fig4}
\end{figure}

To evaluate whether inpainting altered the visual characteristics of the radiographs, we analyzed pixel intensity distributions across the original and cleaned images. As shown in Figure~\ref{fig:fig4}, inpainted images exhibited a more peaked and left-shifted histogram, with grayscale values concentrated in the lower intensity range. Quantitatively, the mean pixel intensity standard deviation was 23.05 ($\pm$ 14.11) for the original images and 31.91 ($\pm$ 9.92) for the inpainted images. Inpainted images exhibit a narrower, left-shifted distribution, indicating a reduction in contrast and grayscale variability.

\section{Discussion}

Generative inpainting using foundation models offers an accessible approach for image editing in medical datasets, including tasks such as de-identification, augmentation, or artifact removal. However, our findings highlight a critical concern: visually plausible edits do not guarantee diagnostic fidelity. Even minor, targeted inpainting, restricted to non-anatomical regions, led to significant degradation in downstream model performance, including both bone age estimation and gender classification.
This performance drop persisted even after bias calibration, suggesting that the edits introduced structural or textural artifacts that altered model-relevant features. Importantly, the gender classification model not only declined in accuracy but also showed increased false male predictions, raising the possibility of inpainting altered features in a demographically biased direction. These results highlight a central concern: generative realism can mask anatomical errors, misleading both clinicians and AI models, especially given that even top-performing bone age models have shown demographic bias in real-world use \cite{beheshtian2023bias}. Such distortions exemplify broader concerns in radiology AI, where bias arises across the full development pipeline, from dataset curation to deployment infrastructure \cite{gichoya2023pitfalls}.
Inpainted images showed altered pixel intensity distributions, with a narrower, left-shifted histogram and lower standard deviation. These shifts reflect changes in grayscale characteristics that may have impacted downstream model performance.
The absence of preprocessing prior to inpainting, while potentially introducing input variability, offered a more realistic test of the model’s robustness. This choice reflects how foundation models are often applied in practice and highlights the need to evaluate their reliability under minimal curation, especially when used as automated preprocessing tools in clinical pipelines.
Some limitations warrant consideration. The inpainting process lacked control over anatomical fidelity, and despite prompts to restrict edits to non-diagnostic regions, changes such as fused growth plates or early ossification centers were introduced, consistently skewing age predictions. This observation aligns with prior work demonstrating that deep learning models rely on specific anatomical landmarks, particularly the carpus and metacarpophalangeal joints, for accurate bone age estimation and that perturbations in these regions can lead to degraded model performance and interpretability \cite{kitamura2022cam, wang2021saliency}. Additionally, the PNG-format input images lacked DICOM metadata, spatial calibration, and modality parameters, limiting the interpretability of pixel-level comparisons. The dataset’s demographic and institutional details, apart from sex, were also undisclosed, precluding stratified analyses of model bias.
Despite these limitations, the implications generalize broadly. In clinical AI tasks that depend on subtle morphological cues, any image manipulation tool that modifies the input distribution can become a source of latent bias or instability. This risk is particularly acute in pediatric imaging, where developmental anatomy underpins clinical inference. As highlighted by Yi et al. \cite{yi2025bias}, rigorous and systematic evaluation of algorithmic bias is essential, especially when modifications to the input data can differentially impact model performance across demographic subgroups. Without such validation, seemingly benign preprocessing methods may undermine fairness, generalizability, or clinical trustworthiness.
Future research should focus on foundation models trained directly on medical images, capable of performing anatomically consistent edits in a fully mask-free manner. Domain-adapted architectures have shown that training on diverse radiologic and other medical imaging data can embed structural and modality-specific priors, enabling more faithful image modification without handcrafted constraints \cite{Paschali_2025}.

\section*{Funding}
This study received no specific funding.

\section*{Data Availability}
The datasets generated and analysed during the current study are available at the following private Figshare link for peer review: 
\url{https://www.kaggle.com/datasets/felipematsuoka/synthetic-hand-x-ray-dataset-for-bone-age/}. 
These data will be made publicly available under a CC BY 4.0 license.

\section*{Code Availability}
The underlying code for this study, including the image generation pipeline, deep learning model training scripts, and statistical analysis tools, is available at the following private Figshare link:
\url{https://github.com/felipe-matsuoka123/EVALUATING-THE-CLINICAL-IMPACT-OF-GENERATIVE-INPAINTING-ON-BONE-AGE-ESTIMATION}.

\newpage

\bibliographystyle{unsrt}  
\bibliography{references}

@misc{openai_gpt_image_1,
  author       = {OpenAI},
  title        = {gpt-image-1},
  howpublished = {\url{https://openai.com/index/image-generation-api/}},
  note         = {Accessed: 2024-07-01}
}

@article{roledeeplearninginmedicalimage,
author = {Santos, Joana Cristo and Alexandre, Hugo and Santos, Miriam and Henriques Abreu, Pedro},
year = {2025},
month = {01},
pages = {},
title = {The Role of Deep Learning in Medical Image Inpainting: A Systematic Review},
volume = {6},
journal = {ACM Transactions on Computing for Healthcare},
doi = {10.1145/3712710}
}

@misc{lugmayr2022repaintinpaintingusingdenoising,
      title={RePaint: Inpainting using Denoising Diffusion Probabilistic Models}, 
      author={Andreas Lugmayr and Martin Danelljan and Andres Romero and Fisher Yu and Radu Timofte and Luc Van Gool},
      year={2022},
      eprint={2201.09865},
      archivePrefix={arXiv},
      primaryClass={cs.CV},
      url={https://arxiv.org/abs/2201.09865}, 
}

@misc{rouzrokh2023multitaskbraintumorinpainting,
      title={Multitask Brain Tumor Inpainting with Diffusion Models: A Methodological Report}, 
      author={Pouria Rouzrokh and Bardia Khosravi and Shahriar Faghani and Mana Moassefi and Sanaz Vahdati and Bradley J. Erickson},
      year={2023},
      eprint={2210.12113},
      archivePrefix={arXiv},
      primaryClass={eess.IV},
      url={https://arxiv.org/abs/2210.12113}, 
}

@article{article,
author = {M B, Jennyfer and Subashini, Parthasarathy},
year = {2023},
month = {01},
pages = {},
title = {Deep Learning Inpainting Model on Digital and Medical Images-A Review},
volume = {20},
journal = {The International Arab Journal of Information Technology},
doi = {10.34028/iajit/20/6/9}
}

@misc{khan2025comprehensivesurveyfoundationmodels,
      title={A Comprehensive Survey of Foundation Models in Medicine}, 
      author={Wasif Khan and Seowung Leem and Kyle B. See and Joshua K. Wong and Shaoting Zhang and Ruogu Fang},
      year={2025},
      eprint={2406.10729},
      archivePrefix={arXiv},
      primaryClass={cs.LG},
      url={https://arxiv.org/abs/2406.10729}, 
}

@article{halabi2019rsna,
  author    = {Halabi, S. S. and Prevedello, L. M. and Kalpathy-Cramer, J. and Mamonov, A. B. and Bilbily, A. and Cicero, M. and Pan, I. and Pereira, L. A. and Sousa, R. T. and Abdala, N. and Kitamura, F. C. and Thodberg, H. H. and Chen, L. and Shih, G. and Andriole, K. and Kohli, M. D. and Erickson, B. J. and Flanders, A. E.},
  title     = {The RSNA Pediatric Bone Age Machine Learning Challenge},
  journal   = {Radiology},
  volume    = {290},
  number    = {2},
  pages     = {498--503},
  year      = {2019},
  month     = feb,
  doi       = {10.1148/radiol.2018180736},
  pmid      = {30480490},
  pmcid     = {PMC6358027}
}

@INPROCEEDINGS{7780459,
  author={He, Kaiming and Zhang, Xiangyu and Ren, Shaoqing and Sun, Jian},
  booktitle={2016 IEEE Conference on Computer Vision and Pattern Recognition (CVPR)}, 
  title={Deep Residual Learning for Image Recognition}, 
  year={2016},
  volume={},
  number={},
  pages={770-778},
  doi={10.1109/CVPR.2016.90}}

@INPROCEEDINGS{5206848,
  author={Deng, Jia and Dong, Wei and Socher, Richard and Li, Li-Jia and Kai Li and Li Fei-Fei},
  booktitle={2009 IEEE Conference on Computer Vision and Pattern Recognition}, 
  title={ImageNet: A large-scale hierarchical image database}, 
  year={2009},
  volume={},
  number={},
  pages={248-255},
  doi={10.1109/CVPR.2009.5206848}}

@misc{cardoso2022monaiopensourceframeworkdeep,
      title={MONAI: An open-source framework for deep learning in healthcare}, 
      author={M. Jorge Cardoso and Wenqi Li and Richard Brown and Nic Ma and Eric Kerfoot and Yiheng Wang and Benjamin Murrey and Andriy Myronenko and Can Zhao and Dong Yang and Vishwesh Nath and Yufan He and Ziyue Xu and Ali Hatamizadeh and Andriy Myronenko and Wentao Zhu and Yun Liu and Mingxin Zheng and Yucheng Tang and Isaac Yang and Michael Zephyr and Behrooz Hashemian and Sachidanand Alle and Mohammad Zalbagi Darestani and Charlie Budd and Marc Modat and Tom Vercauteren and Guotai Wang and Yiwen Li and Yipeng Hu and Yunguan Fu and Benjamin Gorman and Hans Johnson and Brad Genereaux and Barbaros S. Erdal and Vikash Gupta and Andres Diaz-Pinto and Andre Dourson and Lena Maier-Hein and Paul F. Jaeger and Michael Baumgartner and Jayashree Kalpathy-Cramer and Mona Flores and Justin Kirby and Lee A. D. Cooper and Holger R. Roth and Daguang Xu and David Bericat and Ralf Floca and S. Kevin Zhou and Haris Shuaib and Keyvan Farahani and Klaus H. Maier-Hein and Stephen Aylward and Prerna Dogra and Sebastien Ourselin and Andrew Feng},
      year={2022},
      eprint={2211.02701},
      archivePrefix={arXiv},
      primaryClass={cs.LG},
      url={https://arxiv.org/abs/2211.02701}, 
}

@misc{kingma2017adammethodstochasticoptimization,
      title={Adam: A Method for Stochastic Optimization}, 
      author={Diederik P. Kingma and Jimmy Ba},
      year={2017},
      eprint={1412.6980},
      archivePrefix={arXiv},
      primaryClass={cs.LG},
      url={https://arxiv.org/abs/1412.6980}, 
}

@misc{gokcesu2021generalizedhuberlossrobust,
      title={Generalized Huber Loss for Robust Learning and its Efficient Minimization for a Robust Statistics}, 
      author={Kaan Gokcesu and Hakan Gokcesu},
      year={2021},
      eprint={2108.12627},
      archivePrefix={arXiv},
      primaryClass={stat.ML},
      url={https://arxiv.org/abs/2108.12627}, 
}

@article{beheshtian2023bias,
  author    = {Beheshtian, E. and Putman, K. and Santomartino, S. M. and Parekh, V. S. and Yi, P. H.},
  title     = {Generalizability and Bias in a Deep Learning Pediatric Bone Age Prediction Model Using Hand Radiographs},
  journal   = {Radiology},
  volume    = {306},
  number    = {2},
  pages     = {e220505},
  year      = {2023},
  month     = feb,
  doi       = {10.1148/radiol.220505},
  pmid      = {36165796}
}

@article{gichoya2023pitfalls,
  author    = {Gichoya, J. W. and Thomas, K. and Celi, L. A. and Safdar, N. and Banerjee, I. and Banja, J. D. and Seyyed-Kalantari, L. and Trivedi, H. and Purkayastha, S.},
  title     = {AI Pitfalls and What Not To Do: Mitigating Bias in AI},
  journal   = {British Journal of Radiology},
  volume    = {96},
  number    = {1150},
  pages     = {20230023},
  year      = {2023},
  month     = oct,
  doi       = {10.1259/bjr.20230023},
  pmid      = {37698583},
  pmcid     = {PMC10546443}
}

@article{kitamura2022cam,
  author    = {Kitamura, F. C. and Pan, I.},
  title     = {Artificial Intelligence Class Activation Mapping of Bone Age},
  journal   = {Radiology},
  volume    = {303},
  number    = {1},
  pages     = {52--53},
  year      = {2022},
  doi       = {10.1148/radiol.211790}
}

@article{wang2021saliency,
  author    = {Wang, Z. J.},
  title     = {Probing an AI Regression Model for Hand Bone Age Determination Using Gradient-Based Saliency Mapping},
  journal   = {Scientific Reports},
  volume    = {11},
  number    = {1},
  pages     = {10610},
  year      = {2021},
  doi       = {10.1038/s41598-021-90157-y},
  url       = {https://doi.org/10.1038/s41598-021-90157-y},
  issn      = {2045-2322}
}

@article{yi2025bias,
  author    = {Yi, P. H. and Bachina, P. and Bharti, B. and others},
  title     = {Pitfalls and Best Practices in Evaluation of AI Algorithmic Biases in Radiology},
  journal   = {Radiology},
  volume    = {315},
  number    = {2},
  pages     = {e241674},
  year      = {2025},
  doi       = {10.1148/radiol.241674}
}

@article{Paschali_2025,
   title={Foundation Models in Radiology: What, How, Why, and Why                     Not},
   volume={314},
   ISSN={1527-1315},
   url={http://dx.doi.org/10.1148/radiol.240597},
   DOI={10.1148/radiol.240597},
   number={2},
   journal={Radiology},
   publisher={Radiological Society of North America (RSNA)},
   author={Paschali, Magdalini and Chen, Zhihong and Blankemeier, Louis and Varma, Maya and Youssef, Alaa and Bluethgen, Christian and Langlotz, Curtis and Gatidis, Sergios and Chaudhari, Akshay},
   year={2025},
   month=feb }

\end{document}